\documentclass[twocolumn]{el-author}

\newcommand{\ua}{\uparrow}
\newcommand{\nc}{\newcommand}
\nc{\da}{\downarrow} \nc{\hc}{\hat{c}} \nc{\hS}{\hat{S}}
\nc{\bra}{\langle} \nc{\ket}{\rangle} \nc{\eq}{equation (\ref}
\nc{\h}{\hat} \nc{\hT}{\h{T}}\nc{\be}{\begin{eqnarray}}
\nc{\ee}{\end{eqnarray}}\nc{\rd}{\textrm{d}}\nc{\e}{eqnarray}\nc{\hR}{\hat{R}}\nc{\Tr}{\mathrm{Tr}}
\nc{\tS}{\tilde{S}}\nc{\tr}{\mathrm{tr}}\nc{\8}{\infty}\nc{\lgs}{\bra\ua,\phi|}\nc{\rgs}{|\ua,\phi\ket}
\nc{\hU}{\hat{U}}\nc{\lfs}{\bra\phi|}\nc{\rfs}{|\phi\ket}\nc{\hZ}{\hat{Z}}\nc{\hd}{\hat{d}}\nc{\mD}{\mathcal{D}}
\nc{\bd}{\bar{d}}\nc{\bc}{\bar{c}}\nc{\mc}{\mathcal}\nc{\ea}{eqnarray}\nc{\mG}{\mathcal{G}}\nc{\bce}{\begin{center}}
\nc{\ece}{\end{center}}
\date{6th July 2021}

\usepackage{threeparttable}
\usepackage{multirow}
\usepackage{graphicx}
\usepackage{booktabs}
\usepackage{caption}
\begin{document}

\title{Deep Learning to Ternary Hash Codes by Continuation}

\author{Mingrui Chen, Weiyu Li and Weizhi Lu}

\abstract{Recently, it has been observed that  $\{0,\pm1\}$-ternary codes which are simply generated from deep features by hard thresholding, tend to outperform $\{-1, 1\}$-binary codes in image retrieval. To obtain better ternary codes, we for the first time propose to jointly learn the features with the codes by appending a smoothed function to the networks. During training, the function could evolve into a non-smoothed ternary function by a continuation method.  The method circumvents the difficulty of directly training discrete  functions and reduces the quantization errors of ternary codes. Experiments show that the generated  codes indeed could achieve higher retrieval accuracy.}

\maketitle

\section{Introduction}

Existing hashing methods mainly adopt binary codes for image retrieval. The codes are generated by binarizing the features learned by data-independent or data-dependent methods \cite{wang2017survey}. Among the methods, the data-driven deep learning methods tend to perform best \cite{xia2014supervised,lin2015deep,zhu2016deep,liu2016deep}, thanks to their powerful capability in generating discriminative features.  However, the codes generated by deep features are not perfect, subjective to  performance ceilings and even performance decline \cite{Lai2015SimultaneousFL,cao2018hashgan}, with the increasing of code dimension. This is mainly because the deep features with increasing dimensions tend to become sparse, and  their smll/ambiguous elements with values close to zero probably cause large quantization errors and degraded feature discrimination \cite{kong2012double,liu2021ternary}, as they are roughly binarized to be bipolar values +1 or -1. To address the issue, it is natural to introduce a third state 'zero' to specially denote the ambiguous elements, thus yielding $\{0,\pm1\}$-ternary codes. This kind of codes has recently been proved better  than binary codes.

To the best of our knowledge, there are only two methods \cite{Fan2020DeepPN,liu2021ternary} that  have been proposed to generate ternary codes for image retrieval. However, the two methods are suboptimal because they have  features learned and ternarized in two separated steps. Specifically, in \cite{Fan2020DeepPN} the features are generated with a hinge-like loss trained AlexNet \cite{AlexNet2012}, and then ternarized  by two thresholds selected empirically. To obtain better thresholds,  a searching algorithm is proposed in \cite{liu2021ternary}, based on the principle of  maximizing the expectation of pairwise ternary hamming distances between similar samples and decreasing the distances between dissimilar samples.  No matter how well the thresholds are selected,  the  features previously learned alone cannot guarantee to be optimal for the latter ternarization.

To alleviate the issue, we are motivated to jointly learn the features and ternary codes. An intuitive idea is to append a ternary function to the feature extraction network  and then take them as a whole to train. For this framework, the main challenge comes from the optimization of  ternary function, which has zero gradients and makes back-propagation infeasible. To avoid directly training the discrete function, inspired by \cite{cao2017hashnet},  we propose to replace the function with a smoothed function, which could gradually evolve into a desired ternary function during training. This continuation method is known for optimizing discrete functions with guaranteed convergence \cite{allgower2012numerical}. Experiments show that the proposed method indeed outperforms existing ternary hashing methods.

\section{Method} 
\begin{figure}[!t]
\centering
  {\includegraphics[width=0.5\textwidth]{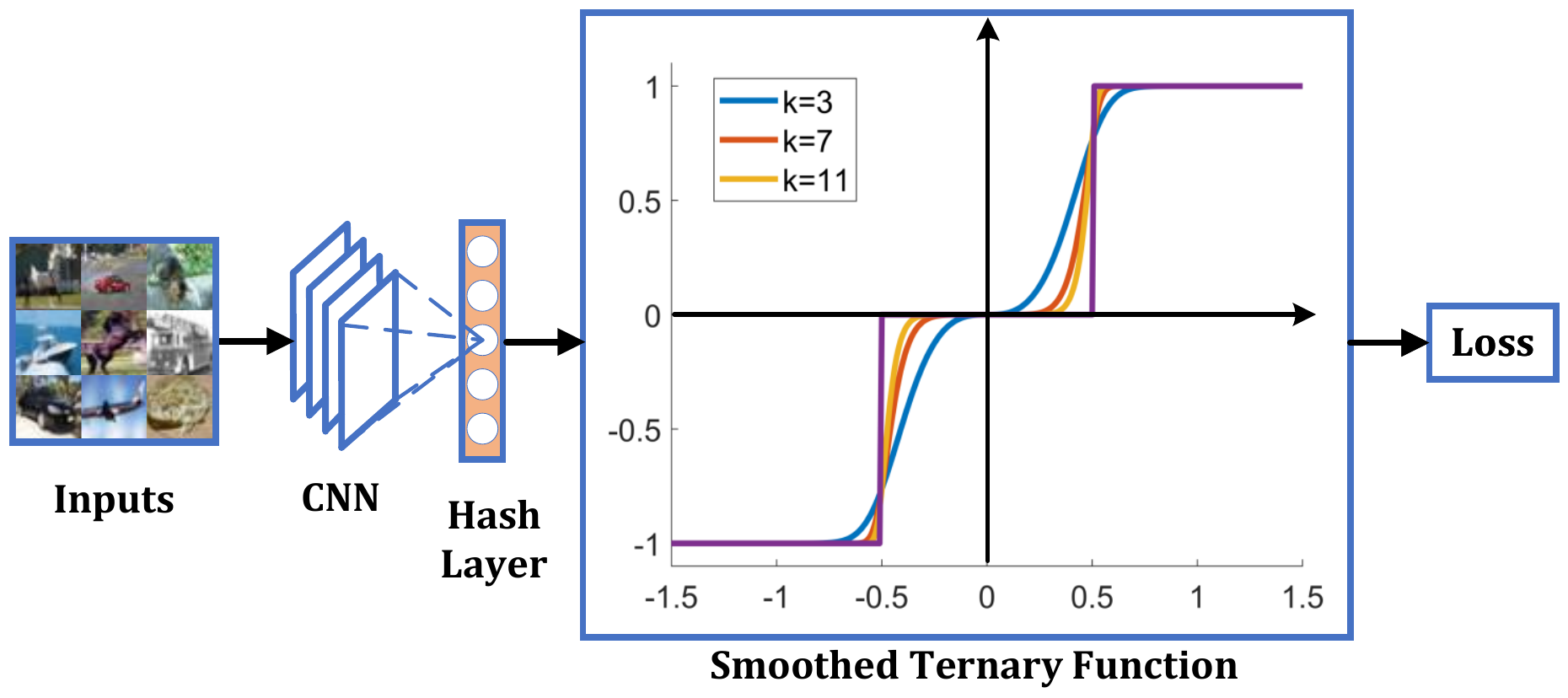}}
\caption{The proposed network architecture for jointly training the features and ternary codes in an end-to-end manner.}
\label{fig_method}
\end{figure}

As illustrated in Figure \ref{fig_method}, the joint learning forms a network pipeline consisting of four  parts: 1) a convolutional neural network (CNN) for learning deep features, 2) a fully connected hash layer  for transforming the features into $d$ dimensions, 3) a smoothed ternary function  for  converting each element of $d$-dimensional features  to be a value close to 1, -1 or 0, and 4) a loss function.  As in \cite{Fan2020DeepPN,liu2021ternary},  we will use AlexNet  for feature learning. To obtain convincing results, we suggest to use the common cross-entropy loss for network training, although other more selective losses, such as hinge-like loss \cite{Fan2020DeepPN}, may lead to better codes.  The smoothed ternary function is proposed as follows:

\begin{equation}  \label{eq_smoothT}
f(x) = tanh({(x/\alpha)}^{k})
\end{equation}
where $\alpha$ is a positive constant and $k$ needs to be an \emph{odd} number greater than one, namely $k=3,5,7$, and so on. The odd $k$ enables $f(x)$ to take both positive and negative values, while  the even $k$ could only lead to non-negative value.  As the odd $k$ tends to infinity, as shown in Figure \ref{fig_method}, $f(x)$ will converge to a ternary function

\begin{equation} \label{eq_hardT}
g(x)=\left\{
\begin{aligned}
1,&&{x      \geq      \alpha}\\
-1,&&{x      \leq      -\alpha} \\
0, &&\text{others}
\end{aligned}
\right.
\end{equation}
where the threshold parameter $\alpha$  is identical to the scale parameter $\alpha$ in \eqref{eq_smoothT}, which is set to be $\alpha=0.5$ in our experiments in terms of the fact that the CNN feature elements $x$ usually have  values normalized to the range $(-1,+1)$.   Considering the equivalence between \eqref{eq_smoothT} and \eqref{eq_hardT} in the limit of $k$, we propose to gradually increase the value of $k$ during training, such that the  smoothed function \eqref{eq_smoothT}  finally approaches the desired ternary function \eqref{eq_hardT}. In testing phase, we need to replace  \eqref{eq_smoothT} with \eqref{eq_hardT} to generate ternary codes, and further, as in  \cite{liu2021ternary}, the ternary codes will be  converted to binary bits via the mapping  $\{-1,0,1\}\rightarrow\{01,00,10\}$,  in order to perform  hamming distances-based image retrieval on binary machines.

\section{Experiments}
We compare the proposed method with other two ternary hashing methods DPN \cite{Fan2020DeepPN}  and TH \cite{liu2021ternary}, on three typical databases CIFAR10 \cite{Krizhevsky09Cifar10}, NUS-WIDE \cite{chua2009nus} and ImageNet100 \cite{ImageNet09}. Following the settings in DHP and TH, the databases are processed as follows: 1) CIFAR10  consists of 60,000  colored images in 10 classes. We randomly select 1000 images (100 images per class) as a query set, and the remaining 59,000 images are taken as a retrieval set.  From the retrieval set, 5,000 images (500 images per class) are randomly selected for network training; 2) NUS-WIDE comprises 269,648 multi-labeled images in 81 classes. By convention, we select the most frequent 21 classes  for experiments. Among them, we randomly select 100 images per class for query and the rest for  retrieval. From the retrieval set, 500 images per class  are randomly sampled for network training; 3) ImageNet contains 1000 categories of images, including over 1.2M images in the training set and 50k images in the validation set. ImageNet100 collects 100 categories from ImageNet, with the entire validation set  for query and the entire training set for retrieval. From the retrieval set, we randomly select 100 images per category for network training.

For generality, as stated before, we simply adopt the AlexNet and cross-entropy loss for feature learning. The network is trained with stochastic gradient decent with 0.9 momentum.   The learning rate is initialized as $10^{-3}$, with cosine learning rate decay. We set the batch size as 64 and the weight decay parameter as $10^{-4}$.  The parameter $k$ for $f(x)$ is gradually increased from 3 to 11 in a total of 150 epoches,  with a stride of 2 every 30 epoches.

The retrieval accuracy is evaluated with  the mean average precision (mAP). As in  DPN \cite{Fan2020DeepPN} and TH \cite{liu2021ternary},  the mAP is calculated with all retrieval images as returned images for CIFAR-10, with top 5,000 returned images for NUS-WIDE, and with top 1,000 returned images for ImageNet100. For fair comparisons, we directly compare with the best results achieved by  DPN and TH  in their original papers \cite{Fan2020DeepPN,liu2021ternary}. Note in TH, the best results are usually obtained by a hinge-like loss, rather than the cross-entropy loss. The results are provided in Table \ref{tab_map}. It is seen that our method achieves consistent performance gains over other two ternary hashing methods, on three databases with varying code dimensions. The gains range from $0.1\%$ to $1.4\%$. So we can say that the proposed joint learning method indeed outperforms exiting  independent learning methods, because of its advantage in reducing quantization errors.

\section{Conclusion} In the letter, we have proposed a continuation method to jointly learn the features and ternary codes in an end-to-end manner. The core of the method is to introduce a smoothed function to gradually approach a desired ternary function during network training, and this avoids the difficulty of directly optimizing the discrete ternary function. As expected, the proposed joint learning method generates better ternary codes than  existing independent learning methods. For generality,  we simply tested the method on the commonly used  AlexNet, which is  trained with cross-entropy loss. It is believed that better ternary codes could be obtained, if more advanced network structures and loss functions are adopted.

\begin{table}[tp]
	\centering
	\fontsize{9}{9}\selectfont
	\caption{Hashing accuracies in terms of mean average precision (mAP)
	\newline for three different methods on  three typical datasets. The best results are
\newline highlighted in bold. \label{tab_map}}
	\label{tabel1}
	\begin{threeparttable}
		\setlength{\tabcolsep}{3mm}{
		\begin{tabular}{cccccc}
			\toprule[2pt]
			
			\multirow{2}{*}{Method}&
			\multicolumn{5}{c}{CIFAR10@all}\cr
			\cmidrule[0.5pt](lr){2-6}
			&16dim&24dim&32dim&64dim&128dim\cr
			\midrule[0.8pt]
			Ours &\textbf{0.830} & \textbf{0.831}&  \textbf{0.840} & \textbf{0.842} & \textbf{0.843}  \cr
			DPN \cite{Fan2020DeepPN} &0.825 & - & 0.838  &0.830 & 0.829\cr
			TH \cite{liu2021ternary}&0.820 & 0.823  &0.836&  0.836 & 0.825 \cr
			
			\midrule[1.2pt]

			\multirow{2}{*}{Method}&
			\multicolumn{5}{c}{NUS-WIDE@5K}\cr
			\cmidrule[0.5pt](lr){2-6}
			&16dim&24dim&32dim &64dim&128dim\cr
			\midrule[0.8pt]
			Ours &\textbf{0.850} & \textbf{0.855}  &\textbf{0.862}&  \textbf{0.865} & \textbf{0.863}\cr
			DPN  \cite{Fan2020DeepPN}&0.847 & - & 0.859  & 0.863 & 0.862\cr
			TH \cite{liu2021ternary}&0.830 & 0.841  &0.843&  0.846 & 0.840 \cr			
			\midrule[1.2pt]

			\multirow{2}{*}{Method}&
			\multicolumn{5}{c}{ImageNet100@1K}\cr
			\cmidrule[0.5pt](lr){2-6}
			&16dim&24dim&32dim&64dim&128dim\cr
			\midrule[0.8pt]
			Ours &\textbf{0.685} & \textbf{0.720}&  \textbf{0.748} & \textbf{0.763} & \textbf{0.767}\cr
			DPN \cite{Fan2020DeepPN}&0.684 & - & 0.740  &0.756 & 0.756\cr
			TH \cite{liu2021ternary}&0.668 & 0.710  &0.732&  0.754 & 0.762 \cr
			
			\bottomrule[2pt]
		\end{tabular}}
	\end{threeparttable}
\end{table}

\vskip3pt
\ack{This work has been supported by the National Natural Science Foundation of China (Grants Nos. 61801264 and  61991412) and the Fundamental Research Funds of Shandong University (Grant No. 2019HW018).}

\vskip5pt

\noindent Mingrui Chen and Weizhi Lu (School of Control Science and Engineering, Shandong University, 73 Jingshi Road Jinan, P. R. China 250061); Weiyu Li (Zhongtai Securities Institute for Financial Studies, Shandong University, 27 Shanda Nanlu, Jinan, P.R.China 250100).
\vskip3pt

\noindent E-mail: wzlu@sdu.edu.cn

\bibliographystyle{ieeetrans}
\bibliography{refs_20210701.bib}
\end{document}